\pdfoutput=1

\documentclass[11pt]{article}

\usepackage[preprint]{acl}

\usepackage{times}
\usepackage{latexsym}
\usepackage{amsmath}
\usepackage{booktabs}
\usepackage{multirow}
\usepackage{amssymb}

\usepackage[T1]{fontenc}

\usepackage[utf8]{inputenc}

\usepackage{microtype}

\usepackage{inconsolata}

\usepackage{graphicx}
\usepackage{tcolorbox}

\tcbuselibrary{theorems}
\newtcbtheorem
  []
  {lesson}
  {Key Takeaway}
  {%
    theorem name,%
    colback=white,
    colframe=black!90,
    fonttitle=\bfseries,
  }
  {tay}

\usepackage{subfigure}
\usepackage{arydshln}
\usepackage{multirow}
\usepackage{enumitem}
\title{Learning Faster with Better Tokens: Parameter-Efficient Vocabulary Adaptation for Specialized Text Summarization}

\author{
  \textbf{Gunjan Balde\textsuperscript{1}},
  \textbf{Soumyadeep Roy\textsuperscript{2}},
  \textbf{Mainack Mondal\textsuperscript{1}}
  and \textbf{Niloy Ganguly\textsuperscript{1}}
\\
  \textsuperscript{1}Dept. of Computer Science and Engg., IIT Kharagpur, Kharagpur, India \\
  \textsuperscript{2}Dept. of Medicine (Biomedical Informatics),
  Stanford University, Stanford, CA, USA
\\
  \small{
    \textbf{Correspondence:} \href{mailto:balde.gunjan0812@gmail.com}{balde.gunjan0812@gmail.com}
  }
}

\begin{document}
\maketitle
\begingroup\renewcommand\thefootnote{}
\footnotetext{\textbf{This is the author’s version of the manuscript. It is posted here for your personal use. Not for redistribution. To Appear in the the 64th Annual Meeting of the Association for Computational Linguistics, ACL (Mains) 2026.}}
\endgroup
\begin{abstract}
Large language models pretrained on general-domain corpora often exhibit tokenization inefficiencies when applied to specialized domains. Although continual pretraining for domain adaptation partially alleviate performance degradation, it does not resolve the fundamental vocabulary mismatch. To address this gap, we introduce a targeted parameter-efficient domain adaptation approach that combines vocabulary adaptation with pretraining for LLM-based text summarization. Our unified framework augments pretrained tokenizers with domain-specific tokens while selectively replacing under-trained and unreachable tokens to limit parameter growth. We evaluate our approach on Llama-3.1-8B and Qwen2.5-7B across legal and medical summarization tasks on a challenge-oriented evaluation protocol focused on expert-driven text and summaries which typically has higher concentration of \textit{over-fragmented} Out-of-Vocabulary (\textbf{OOV}) words. The vocabulary adaptation algorithm enhances the overall quality of the summarization model by improving semantic similarity between the generated summaries and their references. In addition, the adapted model produces summaries that incorporate more appropriate novel and domain-specific words, leading to improved coherence, relevance, and faithfulness. We further observe that our proposed approach significantly reduce training time by $35-55\%$ over continual pretraining and reduce parameter counts up to $37\%$ w.r.t expansion-only methods. We make the codebase publicly available at {\url{https://github.com/gb-kgp/VocabReplace-Then-Expand}}.

\if{0}
However, the added vocabulary construction depends on the LLM tokenizer (and its vocabulary) and the downstream dataset. This means a new model with a new vocabulary has to be trained (using continual p) separately for each dataset and LLM combination, which is computationally inefficient. In this work, we propose the concept of `intrinsic vocabulary' that will represent the added vocabulary for all downstream tasks or datasets for the target domain. Therefore, irrespective of the downstream dataset

Recent efforts on vocabulary adaptation on medical text summarization for large language models (LLMs) are target task-specific. It forces practitioners to rebuild tokenizers and retrain embeddings for every new dataset. This in turn inflates compute, storage and trainable parameters. We show that a target task‑agnostic vocabulary, constructed once from a diverse continual‑pretraining mixture matches the performance of task-specific token sets. Beyond classical expansion, which raises the total trainable parameters by 5–8\%, we introduce a parameter‑efficient replacement scheme. It swaps out \textit{under-utilized} tokens--untrained, undertrained, under-represented, and unreachable tokens--adding zero parameters. We conduct extensive experimentation across four diverse medical summarization benchmarks of varying difficulty on two models. We establish that task-agnostic vocabulary replacement strategies either outperforms or retains up to 98\% of performance over task-specific ones. Our codebase is made publicly available at.
\fi

\end{abstract}

\section{Introduction}

While large language models (LLMs) have revolutionized natural language processing, adapting generalist models to expert domains remains challenging due to high vocabulary mismatch between general and domain-specific corpora. Recent domain-specific models including Meditron-70B~\cite{chen2023meditron70b}; BioMistral~\cite{labrak-etal-2024-biomistral}, built on Mistral-7B and further pretrained on PubMed Central; and PMC-LLaMA~\cite{wu2024pmc} demonstrate that continued pretraining on specialized corpora yields substantial performance improvements. However, vocabulary mismatch fundamentally limits these gains: PubMedBERT~\cite{gu2021domain} demonstrates that medical terms like "naloxone" fragment into meaningless subwords ("nal", "\#\#ox", "\#\#one"), while domain-specific vocabularies treat them atomically. This tokenization inefficiency imposes substantial costs—non-English and domain-specific text can require up to $13\times$ more tokens than English~\cite{rust-etal-2021-good,ahia-etal-2023-languages,NEURIPS2023_74bb24dc}, directly increasing API costs, latency, and memory requirements. Recent work establishes that this fragmentation reduces effective context window size and impedes learning meaningful representations~\cite{hofmann-etal-2022-embarrassingly,kaplan2025from}.

The conventional approach to addressing vocabulary mismatch involves domain-adaptive pretraining (DAPT), where models undergo continued pretraining on domain-specific corpora~\cite{gururangan-etal-2020-dont}. While effective, this paradigm presents significant practical limitations. BioMistral-7B required 32 A100 GPUs for 20 hours on 3 billion tokens from PubMed Central, while Meditron-70B consumed 128 A100 GPUs for 332 hours processing 46 billion tokens while achieving marginal improvements. For contemporary large language models, ~\citet{hu2022lora} note that full fine-tuning is ``prohibitively expensive'', requiring complete parameter updates and storage of separate model instances per domain. While parameter-efficient methods reduce trainable parameters, they do not address the underlying tokenization inefficiency--\textit{vocabulary mismatch}.

An alternative paradigm that directly addresses vocabulary mismatch is vocabulary adaptation, modifying a pretrained model's tokenizer and embedding layer to incorporate domain-specific vocabulary. Recent works~\cite{sachidananda-etal-2021-efficient,hong-etal-2021-avocado,liu-etal-2023-task,yamaguchi2024can,balde2024ijcai,gao2024ve,balde2025evaluationllmsmedicaltext} establishes this as a resource-efficient path. However, vocabulary expansion introduces computational overhead through parameter growth: adding 10,000 tokens to Llama-3-8B requires approximately 80 million additional parameters (at 4096-dimensional embeddings), representing non-trivial increase in model size and inference cost. ~\citet{land-bartolo-2024-fishing} reveal a critical insight: contemporary language models contain $0.1-1\%$ severely under-trained "glitch tokens"—vocabulary tokens that occupy vocabulary slots but contribute minimally due to insufficient pretraining exposure. This observation suggests efficient vocabulary adaptation is possible by strategically replacing under-trained tokens with domain-specific vocabulary, achieving adaptation benefits with minimal parameter expansion~\cite{purason2025teachingoldtokenizersnew}.

In this work, we propose a vocabulary adaptation method that strategically replaces under-trained and unreachable tokens with domain-specific vocabulary before resorting to expansion, thereby minimizing parameter overhead while enabling effective domain specialization. Our approach operates on Llama-3.1-8B and Qwen2.5-7B and consists of four key steps: (1) we train a BPE tokenizer on domain-specific corpora to identify candidate domain vocabulary, (2) we select the top 10,000 tokens based on frequency and coverage statistics as our vocabulary adaptation budget, (3) we compile a replacement candidate list by identifying under-trained and unreachable tokens using~\citet{land-bartolo-2024-fishing} methodology, and (4) we replace tokens from this candidate list with domain-specific tokens, expanding the vocabulary only when the replacement budget is exhausted. This hybrid replacement-then-expansion strategy enables us to prioritize recycling underutilized vocabulary slots, minimizing net parameter increase while maximizing domain vocabulary coverage. 

Beyond standard benchmarking, we introduce a challenge-oriented evaluation framework that stress-tests model performance under conditions where domain vocabulary knowledge is critical. We restructure the downstream domain-specific corpus to explicitly capture challenging scenarios: test sets with high out-of-vocabulary (OOV) concentrations in either source documents (SD) and reference summaries (RS)--\textit{\textbf{OOV\_SD}} and \textbf{\textit{OOV\_RS}} respectively. We also take a \textbf{\textit{Random}} subset without any restriction on OOV concentration to compare the degree of performance in these challenging scenarios. This targeted evaluation approach allows us to assess how well generalist models handle expert-level summarization tasks where domain-specific terminology is essential, providing a more rigorous test of vocabulary adaptation effectiveness beyond aggregate performance metrics. We evaluate our approach on two specialized domains---\textit{medical and legal literature}---demonstrating that our method achieves competitive or superior performance compared to conventional vocabulary extension while substantially reducing parameter overhead and maintaining inference efficiency.

We hypothesize that the effectiveness of vocabulary adaptation is governed by the severity and location of lexical mismatch between pretrained tokenizers and downstream data. We find that: (i) across challenging scenarios of \textbf{\textit{OOV\_SD}} and \textbf{\textit{OOV\_RS}}, we observe more improvement in former setting over competing baselines. Although margins of gain in slightly higher in \textbf{\textit{OOV\_RS}} ($4.44\%$) than \textbf{\textit{OOV\_SD}} ($4.26\%$); (ii) performance gains are notably higher than the gains observed in \textbf{\textit{Random}} setting ($3.06\%$), validating that gains are higher in higher vocabulary mismatch scenario; (iii) vocabulary adaptation enables models to reach their best-performing checkpoints \textbf{35--55\% earlier} than continual pretraining alone, reducing the training time; (iv) hybrid replacement-then-expansion strategy remains highly parameter-efficient reducing parameters by $12.04\%$ and $37.19\%$ for Llama and Qwen models respectively averaged across both the domains. These results identify tokenization mismatch as a bottleneck in domain adaptation and motivate vocabulary-adaptation strategies as a targeted, data-dependent intervention. We make our codebase publicly available at \url{https://github.com/gb-kgp/VocabReplace-Then-Expand}.

\section{Proposed Methodology (\textsc{VocabAdapt})}
\label{sec:methodology}

\subsection{Background}

Generalist LLMs are pretrained on broad-coverage corpora, resulting in tokenizers optimized for general text distributions. When deployed on specialized domains such as medical text, these tokenizers exhibit systematic over-fragmentation. For instance, the term ``Osteoporosis'' is tokenized as \texttt{[O, ste, opor, osis]} by the Llama tokenizer, splitting into four subwords. This over-fragmentation introduces two primary challenges: first, the model must reconstruct semantic meaning across multiple token positions, increasing computational overhead and representation noise; second, generation becomes error-prone as the model must correctly predict each fragment in sequence, with errors compounding across token boundaries.

The standard solution to this vocabulary mismatch problem involves expanding the model's vocabulary by adding domain-specific tokens. Let $V_{\text{src}}$ denote the source vocabulary of size $|V_{\text{src}}|$ with corresponding embedding matrix $E \in \mathbb{R}^{|V_{\text{src}}| \times d}$ and unembedding matrix $U \in \mathbb{R}^{d \times |V_{\text{src}}|}$, where $d$ represents the model's hidden dimension. Adding $k$ domain-specific tokens to form an expanded vocabulary $V_{\text{exp}} = V_{\text{src}} \cup V_{\text{new}}$ necessitates expanding both embedding and unembedding matrices, introducing $2k \cdot d$ additional parameters. For models with large hidden dimensions and substantial domain vocabularies, this parameter overhead becomes significant, increasing memory footprint and inference cost.

We propose an alternative approach that challenges the necessity of vocabulary expansion. Our central hypothesis is that generalist tokenizers contain a substantial subset of undertrained and unreachable tokens that contribute minimally to model performance. Rather than expanding the vocabulary, we identify these ineffectual tokens and replace them with domain-specific terminology, maintaining constant vocabulary size while addressing fragmentation. When domain requirements exceed the available candidate tokens, we resort to expansion only for the remaining terms, thereby minimizing parameter growth.

\subsection{Identifying Candidate Tokens for Replacement}

Our replacement strategy relies on identifying a candidate set $V_{\text{cand}} \subseteq V_{\text{src}}$ comprising tokens that satisfy two independent criteria: they must be undertrained and unreachable.

The undertrained tokens are identified through 
the methodology of \citet{land-bartolo-2024-fishing} where the L2 norm for each token embedding $e_i$ in the vocabulary is computed, $\|e_i\|_2$, excluding partial utf-8, fallback bytes, and unreachable tokens. Their analysis demonstrates that tokens with embedding norms below a threshold corresponds to vocabulary items that appeared infrequently during pretraining and hence undertrained. 
This token token set is henceforth represented as $V_{\text{undertrained}}$.

The unreachable tokens are identified through a consistency test~\cite{land-bartolo-2024-fishing,purason2025teachingoldtokenizersnew}. A token $t$ is deemed unreachable if decoding\footnote{encoding and decoding here corresponds to buit-in \texttt{tokenizer.encode} and \texttt{tokenizer.decode} function calls of a model tokenizer.} its corresponding vocabulary token-id $t_i$ and encoding the decoded token does not yield the original token-id $t_i$. E.g. decoding the encoding token-id $378$ in Llama-3.1-8B results in \texttt{âĢ}, which upon encoding yield token-id $5809$. Formally, a token is unreachable when $\text{encode}(\text{decode}(t_i)) \neq [t_i]$. These tokens represent vocabulary entries that cannot be produced through the standard tokenization algorithm and thus remain inaccessible during normal model inference. While they occupy vocabulary slots and contribute to parameter count, they serve no functional role in model operation. This token set is henceforth represented as $V_{\text{unreachable}}$. 

We define our candidate set as the union of these two criteria: 
\begin{equation}
V_{\text{cand}} = V_{\text{undertrained}} \cup V_{\text{unreachable}}
\end{equation}
This union ensures we replace tokens that are poorly trained and inaccessible, providing a conservative strategy that minimizes risk of degrading model performance on general domains. Empirically, we observe that approximately $3-4\%$ percent of vocabulary tokens in both Llama-3.1-8B and Qwen2.5-7B satisfy the candidate set criterion, providing a substantial pool of replacement candidates.

We apply a final refinement to ensure tokenizer integrity. BPE (Byte-Pair Encoding) subword tokenization algorithm construct vocabulary through iterative merge operations, where character sequences are progressively combined into larger units based on merge rules. Replacing a token that appears in the merge rule of another token outside the candidate set would fundamentally break the tokenization process, rendering certain vocabulary tokens untokenizable. To prevent this, we filter the candidate set to exclude any token that appears as a component in the merge rule of a token not designated for replacement. We construct a directed acyclic graph (DAG) with nodes as the token-id and an edge from token-i to token-j marking the relationship if token-i contributed in merge-rule of token-j (E.g., in $\rightarrow$ ing). Then, for every candidate that could be replaced, we checked if it has any descendants (nodes reachable from this node) that lies outside the candidate replacement set. If yes, we do not replace it, else we consider it for replacement. This set of tokens is marked as $V_{\text{exclude}}$. This constraint guarantees that all remaining merge rules remain valid after vocabulary modification, preserving the deterministic and complete nature of the tokenization algorithm. The refined candidate set therefore contains only tokens that are undertrained, unreachable, and removing does not compromise the structural integrity of the tokenizer.

\begin{equation}
V_{\text{cand}} = V_{\text{cand}} \backslash V_{\text{exclude}}
\label{eq:final_candidate}
\end{equation} 

The final replacement candidate set is of size $1528$ for Llama-3.1-8B (vocabulary size: 128K) and $3987$ for Qwen-2.5-7B (vocabulary size: 151K). We next describe our domain-specific vocabulary construction step.

\subsection{Building Domain-Specific Vocabulary}

We construct domain-specific vocabulary through a  process involving corpus curation, independent tokenizer training, and vocabulary filtering for each target domain. This approach ensures that our added tokens genuinely represent domain-salient terminology rather than arbitrary subword fragments.

We curate two domain-specific corpora, each comprising 100 million tokens (100M) sampled from authoritative sources within their respective domains. The medical domain corpus is sampled from the MEDITRON pretraining corpora \citep{chen2023meditron70b}, which aggregates clinical practice guidelines, PubMed Central full-text articles, and article abstracts, providing comprehensive coverage of both clinical and biomedical language. For the legal domain, we compile a corpus from Supreme Court of India case documents, capturing the specialized vocabulary and linguistic conventions of Indian jurisprudence. 

We train an independent Byte-Pair Encoding tokenizer using the HuggingFace tokenizers~\footnote{\url{https://github.com/huggingface/tokenizers}} library with a vocabulary size of 256,000 tokens dor each domain corpus. This training process learns domain-optimized merge operations that naturally surface frequently occurring domain-specific terms as single tokens. From each trained domain tokenizer vocabulary, we extract candidate tokens for addition to the base model. We filter this set to exclude any tokens that already exist in the source model vocabulary $V_{\text{src}}$, as these tokens require no adaptation. This non-overlapping constraint ensures we only add genuinely new vocabulary items that address coverage gaps in the original tokenizer.

We apply an additional refinement to ensure linguistic coherence across models and avoid introducing problematic tokens. We restrict the candidate set to tokens containing only English alphabetic characters, excluding any subwords that contain numeric digits, special symbols, or mixed alphanumeric patterns. This filtering serves multiple purposes: it eliminates formatting artifacts, date fragments, and identifier components that do not represent meaningful linguistic units; it ensures that added tokens correspond to genuine lexical items rather than incidental character sequences; and it maintains consistency with the predominantly alphabetic nature of established vocabulary in pretrained models. The resulting filtered set forms our domain-specific vocabulary $V_{\text{new}}^{\mathcal{D}}$, comprising high-frequency, domain-salient, purely alphabetic tokens that address the most significant tokenization inefficiencies for the target domain. 

In both the settings, we select the top 10,000 vocabulary tokens ranked by frequency in the domain corpus, representing the most salient domain-specific vocabulary items. We next describe the procedure of vocabulary replacement.

\subsection{Vocabulary Replacement-Then-Expansion and Embedding Initialization}

Thus far, we have a domain vocabulary $V_{\text{new}}^{\mathcal{D}}$ and replacement candidate set $V_{\text{cand}}$(Eq.~\ref{eq:final_candidate}), such that $|V_{\text{new}}^{\mathcal{D}}| > |V_{\text{cand}}|$ . We first replace the $V_{\text{cand}}$ from LLM's base vocabulary with equal sized set from $V_{\text{new}}^{\mathcal{D}}$ sorted by the natural merge order. We then expand the base vocabulary with the remaining $|V_{\text{new}}^{\mathcal{D}}| - |V_{\text{cand}}|$ elements from $V_{\text{new}}^{\mathcal{D}}$. 

Initializing embeddings for the newly replaced and added tokens presents a critical challenge, as random initialization would require substantial training to achieve reasonable representations. Instead, we employ subword aggregation~\cite{yamaguchi2024can}, leveraging model's existing understanding of subwords. For each new token $t_{\text{new}}$, we tokenize it using the original tokenizer to obtain a sequence of source tokens $[t_1, \ldots, t_n]$. We then initialize the new token's embedding as the mean of these constituent embeddings: 
\begin{equation}
e_{t_{\text{new}}} = \frac{1}{n}\sum_{i=1}^{n} e_{t_i}
\end{equation}
This initialization provides a reasonable starting point that captures compositional semantics while allowing subsequent training to refine the representation. The same subword aggregation strategy is applied to initialize the corresponding unembedding matrix row. Next, we describe the procedure to tune the model with the modified vocabulary.


\subsection{Domain-Specific Continual Pretraining}

Following vocabulary modification, we conduct domain-specific continual pretraining to adapt the model to the target domain while training the new token representations. We employ Low-Rank Adaptation (LoRA) \citep{hu2022lora} to enable parameter-efficient training, inserting trainable low-rank matrices into the model's attention and feed-forward layers while keeping the original pretrained parameters frozen. This approach substantially reduces the number of trainable parameters and memory requirements during adaptation.

Each domain model is trained independently on a domain-specific corpus of 100M tokens sampled from high-quality sources representative as discussed previously. We train using the standard causal language modeling objective with next-token prediction, optimizing the model to predict each token given all preceding context. Training is conducted separately for medical and legal domains, producing two specialized model variants from each base model architecture.
\section{Experimental Setup}
Here, we describe the evaluation metrics and datasets used, followed by the baseline models and implementation details. 

\begin{table*}[t]
\centering
\scriptsize
\setlength{\tabcolsep}{0.2cm}
\begin{tabular}{c|c|cc|cc|cc|cc}
\hline
 & \multicolumn{1}{c|}{\textbf{Corpus}} & \multicolumn{2}{c|}{\textbf{SD Token Count}} & \multicolumn{2}{c|}{\textbf{RS Token Count}} & \multicolumn{2}{c|}{\textbf{SD OOV Conc.}} & \multicolumn{2}{c}{\textbf{RS OOV Conc.}} \\
 & \multicolumn{1}{c|}{\textbf{Size}} & \multicolumn{1}{c}{\textbf{Llama}} & \multicolumn{1}{c|}{\textbf{Qwen}} & \multicolumn{1}{c}{\textbf{Llama}} & \multicolumn{1}{c|}{\textbf{Qwen}} & \multicolumn{1}{c}{\textbf{Llama}} & \multicolumn{1}{c|}{\textbf{Qwen}} & \multicolumn{1}{c}{\textbf{Llama}} & \multicolumn{1}{c}{\textbf{Qwen}} \\ \hline
\multicolumn{10}{c}{\textbf{Medical}} \\
\textbf{Random} & 399 & 823 &	847 & 150 & 153 & 11.83 & 11.88 & 13.51 & 13.59 \\ 
\textbf{\textit{OOV\_SD}} & 399 & 808 & 828 & 147 & 150 & 16.90 & 16.94 & 17.23 & 17.24 \\ 
\textbf{\textit{OOV\_RS}} & 399 & 837 & 862 & 137 & 139 & 14.40 & 14.43 & 21.61 & 21.65 \\ \hline
\multicolumn{10}{c}{\textbf{Legal}} \\
\textbf{Random} & 711 & 5870 & 6059 & 1171 & 1221 & 4.75 & 4.76 & 4.76 & 4.76 \\ 
\textbf{\textit{OOV\_SD}} & 710 & 4661 & 4801 & 940 & 975 & 7.48 & 7.49 & 6.97 & 6.98 \\ 
\textbf{\textit{OOV\_RS}} & 711 & 5058 & 5214 & 856 & 889 & 6.39 & 6.40 & 8.57 & 8.57 \\ \hline

\end{tabular}
\caption{Dataset statistics across Legal and Medical domains under \textbf{\textit{Random}}, \textbf{\textit{OOV\_RS}}, and \textbf{\textit{OOV\_SD}} settings, reporting mean token counts, OOV concentration (fraction of unigrams in text split more than once), and novel unigram concentration (fraction of unigrams in RS not present in SD). Medical domain exhibits higher OOV concentrations than Legal domain  Legal domain has substantially higher token counts than Medical domain.} 
\label{tab:dataset_desc} 
\end{table*}

\paragraph{Datasets.} We test our pipeline on two summarization datasets one from each domain. We use the English subset of MultiClinSumm dataset~\cite{Lima_Lopez2025-se} for medical domain. The dataset comprises clinical case reports as source document (SD) and their corresponding summaries derived from case report as the reference summaries (RS). We use the abstractive summarization dataset (IN-ABS) proposed in ~\citet{shukla-etal-2022-legal} for Legal domain. Here SD is a court case judgment from an Indian court and RS is an abstractive summary of the case judgment. To understand the generalizability of our approach across tasks, we further supplement the evaluation for medical domain on two summarization tasks: Evidence-based summarization~\cite{molla2011development} and patient healthcare query summarization~\cite{ben-abacha-demner-fushman-2019-summarization,van2024adapted}. The EBM (Evidence-based Summarization) comprises a query accompanied by a PubMed abstract as a context as the source document and the reference summary as answer to the question in context of the query. CHQ (Patient healthcare query summarization) consists of the a patient-written healthcare query as input and a medical-expert written one-line concise question for the patient query as the summary. In the main text we discuss the results using clinical report summarization and the results for EBM and CHQ datasets in Appendix~\ref{sec:appendix_expt_results}. 

\paragraph{Restructuring Datasets for Expert-Level Summaries.} We restructure the standard dataset in such a way that challenging data points constitute our test set~\cite{balde2024ijcai,balde2025evaluationllmsmedicaltext}. We specifically consider two scenarios where: a) the source documents have higher OOV concentration--\textit{\textbf{OOV\_SD}}, b) the reference summaries have higher OOV (Out-of-Vocabulary) concentration--\textit{\textbf{OOV\_RS}}. The top-10\% of data points from each of these categories  are considered higher concentration documents which constitute our restructured test set. The rest 90\% of corpus is kept as training set. Additionally, we create an equal-sized \textbf{\textit{Random}} train/test subset without any restrictions on OOV concentrations to understand the degree of improvements in challenging scenarios. The dataset statistics are reported in Table~\ref{tab:dataset_desc}. We note that there is roughly $30-40\%$ overlap in the test set of challenging scenarios. 

\begin{table}
    \centering
    \scriptsize
    \begin{tabular}{p{7cm}}
        \hline
        \multicolumn{1}{c}{\textbf{Prompt structure}} \\ \hline

        \multicolumn{1}{c}{\textbf{Medical}} \\
            You are an expert medical professional. \\
            \#\#\#\# \\
            Summarize the given clinical case report into a discharge summary of 100 words or less. Use the examples to guide word choice. \\

            Clinical Case Report 1: \{Train-Case-Document\} \\
            Discharge Summary 1 : \{Train-Summary\} \\
            \#\# \\

            Clinical Case Report 2: \{Test-Case-Document\} \\
            Discharge Summary 2 : \\ \hline      

        \multicolumn{1}{c}{\textbf{Legal}} \\
            You are an expert Indian Legal professional. \\
            \#\#\#\# \\
            Summarize the given legal case document in 300 words on less. Use the examples to guide word choice. \\

            Case Document 1: \{Train-SD\} \\
            Summary 1 : \{Train-RS\} \\
            \#\# \\

            Case Document 2: \{Test-SD\} \\
            Summary 2 : \\ \hline

    \end{tabular}
    \caption{The prompt structure used for prompting LLMs inspired based on the structure proposed in ClinSumm~\cite{van2024adapted}. Since we are using BASE LLMs there is no explicit segregation of system prompt and user prompt.}
    \label{tab:prompt_structure}
\end{table}

\paragraph{Baseline Models.} We used the base variants of two LLMs - Qwen-2.5~\cite{yang2024qwen2technicalreport} (Model id: \href{https://huggingface.co/Qwen/Qwen2.5-7B}{Qwen/Qwen2.5-7B}), and Llama-3.1~\cite{touvron2023llama} (Model id: \href{https://huggingface.co/meta-llama/Llama-3.1-8B}{meta-llama/Llama-3.1-8B}) as our \textit{BASE} models. They do not undergo vocabulary adaptation and continual pretraining. Additionally, we also used continually pretrained variants of these base models on domain-specific text, which we label as `CPTOnly (No Vocab Adapt)'. This helps us to evaluate the improvements observed solely because of vocabulary adaptation.

\paragraph{Training and Inference Strategy.} All the experiments are conducted on a single H100 80 GB GPU. We train the models using standard causal language modeling task of next token prediction and use greedy decoding to generate summaries. We use LoRA and set rank at 32, alpha at 64, learning rate at $2e-5$. For all the domains, we adapt a vocabulary of size $10K$ and train the models on $100M$ tokens dataset for a total of 3 epochs with an effective batch size of 64. Both CPTOnly and \textsc{VocabAdapt} are trained on identical corpora and hyperparameter setting, with \textsc{VocabAdapt} additionally performing a one-time vocabulary construction step that takes roughly 30 minutes (on a single core of Apple M3 Pro laptop). Despite this overhead, \textsc{VocabAdapt} completes training in $6.5–8.5$ hours total, making it notably faster than CPTOnly, which requires $10.5–12.5$ hours. For inference, we use in-context learning~\cite{brown2020languagemodelsfewshotlearners} to provide inputs to model with only one example demonstration appended to the test data point (Appendix~\ref{sec:appendix_sample_ICL} contains details on the sampling procedure for ICL demonstration). The prompt structure for ICL is provided in Table~\ref{tab:prompt_structure}. 

\paragraph{Evaluation Metrics.} We evaluate the summarization quality using Rouge-LCS (R-LCS) as the main evaluation metric and report F-score values, as followed by prior works~\cite{balde2024ijcai,balde2025evaluationllmsmedicaltext,fabbri-etal-2021-summeval}. We also report BertScore~\cite{2020BERTScore} where we use BioBert~\cite{lee2020biobert} embeddings and InLegalBERT~\cite{cite:InLegalBert} embeddings for the medical and legal domain evaluation respectively. We also conduct a LLM-as-judge evaluation of the summaries generated in medical and legal domains. We use the Google's MedGemma-27B model~\cite{sellergren2025medgemmatechnicalreport} for medical domain and Gemma3-27B~\cite{gemma3technicalreport} for legal domain to evaluate the model-generated summaries across three evaluation dimensions: coherence, relevance, and faithfulness on a scale of $1-5$~\cite{fabbri-etal-2021-summeval,zhang2023famesumm}.
\section{Experimental Results}

\begin{table*}[t]
\centering
\scriptsize
\setlength{\tabcolsep}{0.15cm}
\scalebox{0.9}{
\begin{tabular}{lrr:rr:rr:rr:rr:rr:r}
\hline
 & \multicolumn{1}{c}{Best} & \multicolumn{4}{c}{\textit{\textbf{Random}}} & \multicolumn{4}{c}{\textit{\textbf{OOV\_SD}}} & \multicolumn{4}{c}{\textit{\textbf{OOV\_RS}}} \\ 
 
 & \multicolumn{1}{c}{Ckpts.} & \multicolumn{1}{c:}{FrSr\textsubscript{SD}}$\downarrow$ & \multicolumn{1}{c|}{FrSr\textsubscript{RS}}$\downarrow$ & \multicolumn{1}{c:}{R-LCS}$\uparrow$ & \multicolumn{1}{c|}{BSr}$\uparrow$ & \multicolumn{1}{c:}{FrSr\textsubscript{SD}}$\downarrow$ & \multicolumn{1}{c|}{FrSr\textsubscript{RS}}$\downarrow$ & \multicolumn{1}{c:}{R-LCS}$\uparrow$ & \multicolumn{1}{c|}{BSr}$\uparrow$ & \multicolumn{1}{c:}{FrSr\textsubscript{SD}}$\downarrow$ & \multicolumn{1}{c|}{FrSr\textsubscript{RS}}$\downarrow$ & \multicolumn{1}{c:}{R-LCS}$\uparrow$ & \multicolumn{1}{c}{BSr}$\uparrow$ \\ \hline
\multicolumn{13}{c}{\textbf{Medical}} \\
Llama-3.1-8B-BASE & - & 1.16 & 1.16 & 23.03 & 70.66 & 1.25 & 1.27 & 24.39 & 71.35 & 1.20 & 1.34 & 21.33 & 70.41 \\
CPTOnly (No Vocab Adapt) & 7500  & 1.16 & 1.16 & 24.89 & 75.55 & 1.25 & 1.25 & 26.29 & 76.22 & 1.20 & 1.34 & \textbf{23.92} & 75.43 \\
\textsc{VocabAdapt} & 3500 &\textbf{1.05} & \textbf{1.06} & \textbf{24.98} & \textbf{75.98} & \textbf{1.09} & \textbf{1.09} & \textbf{26.68} & \textbf{76.55} & \textbf{1.06} & \textbf{1.12} & 23.65 & \textbf{75.58} \\ \hdashline
Qwen2.5-7B-BASE & - & 1.19 & 1.23 & 15.51 & 43.35 & 1.28 & 1.29 & 14.15 & 37.44 & 1.24 & 1.36 & 12.23 & 35.22 \\
CPTOnly (No Vocab Adapt) & 8000 & 1.19 & 1.23 & 24.73 & 75.32 & 1.28 & 1.29 & 25.96 & 75.90 & 1.24 & 1.36 & 22.41 & 74.44 \\
\textsc{VocabAdapt} & 3500 & \textbf{1.09} & \textbf{1.10} & \textbf{25.04} & \textbf{75.72} & \textbf{1.12} & \textbf{1.11} & \textbf{26.11} & \textbf{76.15} & \textbf{1.10} & \textbf{1.14} & \textbf{23.12} & \textbf{75.21} \\ \hdashline
\multicolumn{13}{c}{\textbf{Legal}} \\
Llama-3.1-8B-BASE & - & 1.03 & 1.03 & \textbf{25.89} & 67.04 & 1.08 & 1.06 & 24.82 & 64.10 & 1.06 & 1.08 & \textbf{24.86} & 65.28 \\
CPTOnly (No Vocab Adapt) & 10000 & 1.03 & 1.03 & 25.36 & 69.05 & 1.08 & 1.06 & 24.83 & 68.04 & 1.06 & 1.08 & 24.36 & \textbf{68.38} \\
\textsc{VocabAdapt} & 6500 & \textbf{1.01} & \textbf{1.01} & 25.42 & \textbf{69.14} & \textbf{1.01} & \textbf{1.01} & \textbf{24.89} & \textbf{68.12} & \textbf{1.01} & \textbf{1.01} & 23.92 & 68.11 \\  \hdashline

Qwen2.5-7B-BASE & - & 1.06 & 1.06 & 10.63 & 28.25 & 1.11 & 1.10 & 9.97 & 27.64 & 1.09 & 1.12 & 9.90 & 27.15 \\
CPTOnly (No Vocab Adapt) & 10500 & 1.06 & 1.06 & \textbf{25.68} & \textbf{69.04} & 1.11 & 1.10 & \textbf{25.16} & \textbf{67.60} & 1.09 & 1.12 & \textbf{24.72} & \textbf{67.97} \\
\textsc{VocabAdapt} & 6500 & \textbf{1.02} & \textbf{1.02} & 24.23 & 67.89 & \textbf{1.05} & \textbf{1.05} & 23.60 & 66.19 & \textbf{1.04} & \textbf{1.05} & 23.22 & 66.89 \\ \hline

\end{tabular}}
\caption{Comparison of best vocabulary adaptation methods across different domains (Legal and Medical) using in-context learning with one exemplar demonstration in two challenging scenarios --\textbf{\textit{OOV\_RS}} and \textbf{\textit{OOV\_SD}} and \textbf{\textit{Random}} subset. We report Rouge-LCS (\textbf{R-LCS}), BERTScore (\textbf{BSr}), and Fragment scores in SD ({\bf FrSr\textsubscript{SD}}) and RS ({\bf FrSr\textsubscript{RS}}). We note that: (i) vocabulary adaptation significantly brings down fragment scores, (ii) improvement margins are higher in medical domain compared to legal domain (owing to higher OOV concentration), (iii) improvements due to vocabulary adaptation is typically higher in challeneging scenarios than \textbf{\textit{Random}} setting. (iv) vocabulary adaptation brings down training time by $35-55\%$ compared to CPTOnly baselines. This contrast between \textbf{\textit{OOV\_SD}} and \textbf{\textit{OOV\_RS}} highlights that source-side OOV primarily affects content understanding, while reference-side OOV impacts lexical realization, and effective vocabulary adaptation is crucial in addressing both challenges beyond what is observed in the \textbf{\textit{Random}} setting.} 
\label{tab:full-results}
\end{table*}

We report Rouge-LCS, BERTScore, and Fragment Scores (avg. number of subwords a word is tokenized into) in Table~\ref{tab:full-results} focusing on best vocabulary adaptation strategies. Further results are provided in Appendix~\ref{sec:appendix_expt_results}. We observe that the impact of vocabulary expansion is strongly domain-dependent. Improvements are more pronounced in medical domain which has higher OOV concentration as compared to legal domain. We now provide a detailed discussion across scenarios highlighting where vocabulary adaptation does and does not work.

\paragraph{Vocabulary adaptation leads to a lower fragment score.} Vocabulary adaptation techniques improve fragment score, thus reducing over-fragmentation and addressing vocabulary mismatch. In medical domain, we see a reduction of $16.02\%$ and $15.63\%$ for Llama and Qwen respectively across challenging OOV scenarios. In legal domain, we see a reduction of $5.95\%$ and $5.73\%$ for Llama and Qwen respectively across challenging scenarios. This reduction makes models energy-efficient as fewer tokens are needed to encode and generate compared to BASE, resulting in better representations.

\paragraph{Vocabulary Adaptation improves more in OOV concentration subset of source document versus reference summary.} Vocabulary adaptation improves in all cases (in terms of R-LCS and BERTScore) over BASE and 6 out of 8 comparisons over CPTOnly in \textbf{\textit{OOV\_SD}}. In \textbf{\textit{OOV\_RS}}, vocabulary adaptation improves in a total of 7 out of 8 comparisons over BASE and only 3 out of 8 comparisons over CPTOnly. The observed improvement can be attributed to a greater reduction in source-side token fragmentation — $10.16\%$ for \textbf{\textit{OOV\_SD}}  compared to $8.92\%$ for \textbf{\textit{OOV\_RS}}. Higher fragmentation in \textbf{\textit{OOV\_RS}} leads to a more dispersed attention distribution, which can hinder the model’s ability to effectively capture and understand the source document, ultimately affecting overall performance.

\paragraph{Improvements in medical domain is higher than legal domain.} Although for both the domains vocabulary adaptation has consistently improved over BASE. This behavior could be tied to rather simple observation from Table~\ref{tab:dataset_desc}. Medical domain has substantially higher OOV concentrations in source documents and reference summaries which make it an ideal candidate for vocabulary adaptation. 

\paragraph{Improvement in \textit{Random} is moderate compared to \textit{OOV} settings.} \textbf{\textit{Random}} setting yields slightly lower absolute performance than challenging OOV scenarios for medical domain (Qwen: 75.72 vs \textbf{\textit{OOV\_SD}} 76.15 BSr; Llama: 75.98 vs \textbf{\textit{OOV\_SD}} 76.55 BSr), validating that vocabulary adaptation is most beneficial under severe OOV constraints. The performance gap between \textbf{\textit{Random}} and OOV scenarios is more pronounced in medical domain, consistent with higher OOV concentration in SD and RS subsets. Fragmentation reduction is more substantial in OOV scenarios than \textbf{\textit{Random}} for medical domain (Qwen: FrSr 1.10-1.14 in OOV vs 1.09-1.10 in Random), demonstrating \textsc{VocabAdapt} higher performance is consistent with higher reduction in fragmentation. That said, it needs to be mentioned that in a random situation \textsc{VocabAdapt} has positive impact albeit small. 

\begin{figure}[t]
    \centering
    \includegraphics[width=0.45\textwidth]{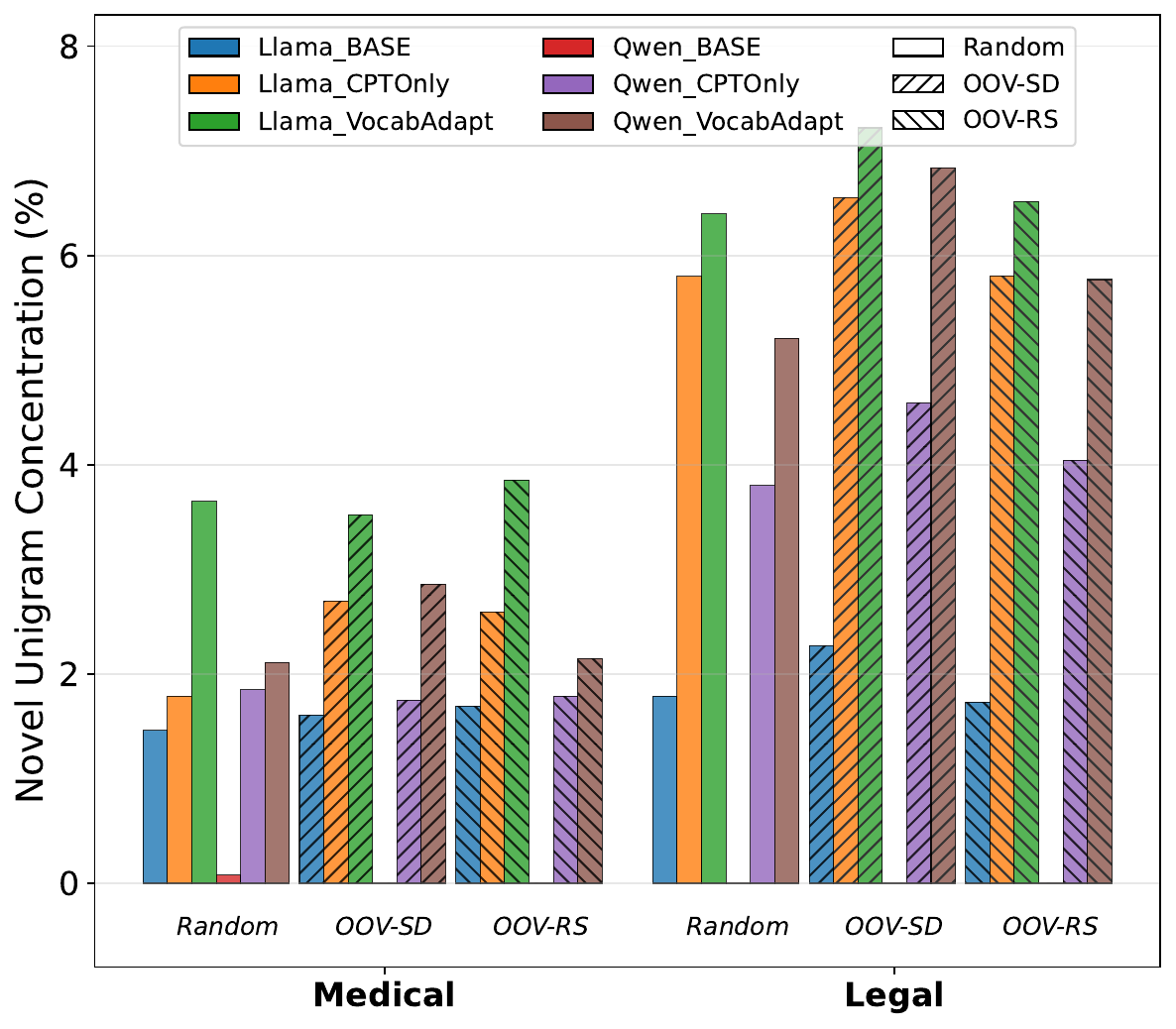}
    \caption{Median novel unigram concentration observed in the summaries generated by BASE, CPTOnly, and \textsc{VocabAdapt} methods for Llama and Qwen models. We note that vocabulary adaptation method brings more (meaningful) novel words compared to baselines.}
    \label{fig:placeholder}
\end{figure}

\paragraph{Vocabulary adaptation improves training efficiency.}
Beyond final performance, vocabulary adaptation methods consistently achieve their best checkpoints substantially earlier than CPTOnly across domains and model families. Concretely, from Table~\ref{tab:full-results}, we note that  vocabulary adaptation variants results in an approximate \textbf{35--55\% reduction in training steps} to peak performance compared to CPTOnly.This in turn reduce training time while maintaining similar performance or even outperforming CPTOnly. This indicates that correcting tokenization mismatch improves optimization efficiency by allowing models to allocate capacity to coherent domain tokens.

\paragraph{Vocabulary adaptation improves semantic overlap.}
We conduct a brief analysis to understand why in certain cases there is a slight drop in Rouge-LCS but gains in BERTScore. We hypothesize that since vocabulary adaptation results in higher abstraction--generation of novel unigrams that are absent in the source document is more prevalent. This might in turn brings terms that are not lexically overlapping with reference summary but does carry similar semantics. We report our findings in Figure~\ref{fig:placeholder}. We note that vocabulary adaptation indeed introduce more meaningful abstraction (novel unigrams) than baselines consistently across evaluation scenarios. Thus validating it might account for slight drop in Rouge-LCS complemented by increase in BERTScore. The next question which can be asked is whether  introduction of such novel words improve the readability, coherence of the summary, we answer that question by using LLM as a judge. 

\paragraph{LLM-as-a-Judge Evaluation for measuring quality of a generated summary.}  We conduct a LLM-as-a-Judge evaluation~\cite{croxford2025automating} of the summaries generated by CPTOnly and vocabulary adaption methods in medical and legal domain. We conduct evaluation across three dimensions of coherence, relevance, and faithfulness as done in prior art~\cite{zhang2023famesumm,balde2024ijcai,balde2025evaluationllmsmedicaltext}. We take 100 random samples from medical domain and 20 from legal domain distributed uniformly across OOV scenarios and models. We report the average scores in Table~\ref{tab:llaj-scores}. We find that vocabulary adaptation generates more coherent, relevant, and faithful summaries compared to competitive CPTOnly baseline. (See Appendix~\ref{sec:appendix_llaj} for further details).
\begin{table}[!ht]
    \centering
    \scriptsize
    \scalebox{0.8}{
    \begin{tabular}{lrrr}
    \hline
         & \multicolumn{1}{c}{\textbf{Coherence}} & \multicolumn{1}{c}{\textbf{Relevance}} & \multicolumn{1}{c}{\textbf{Faithfulness}} \\ \hline
        \multicolumn{4}{c}{\textbf{Medical}} \\
        Llama-CPTOnly (No Vocab Adapt)    &   2.70    &   3.30    &   3.48 \\
        Llama-\textsc{VocabAdapt}                  &   \textbf{3.30}    &   \textbf{3.82}    &   \textbf{3.84} \\ \hdashline
        Qwen-CPTOnly (No Vocab Adapt)    &   2.56    &   3.34    &   3.76 \\
        Qwen-\textsc{VocabAdapt}                  &   \textbf{3.24}    &   \textbf{3.92}    &   \textbf{3.94}   \\ \hline

        \multicolumn{4}{c}{\textbf{Legal}} \\
        Llama-CPTOnly (No Vocab Adapt)    &   2.70 & 2.80 & 2.30 \\
        Llama-\textsc{VocabAdapt}         &   \textbf{2.70} &  \textbf{3.60} & \textbf{3.80} \\ \hdashline
        Qwen-CPTOnly (No Vocab Adapt)     &    \textbf{2.60} & 1.90 & 3.60 \\
        Qwen-\textsc{VocabAdapt}          &   2.20 & \textbf{2.50} & \textbf{4.30}   \\ \hline

    \end{tabular}}
    \caption{LLM-as-a-Judge results for Medical domain using MedGemma-27B, and Gemma-27B model for Legal domin as the evaluator. The evaluation is carried out across coherence, relevance, and faithfulness on a scale of $1-5$. We observe that the summaries generated by vocabulary adaptation methods are mostly rated higher than CPTOnly baseline, resulting in better summaries.}
    \label{tab:llaj-scores}
\end{table}

\begin{table}[h]
\centering
\scriptsize
\setlength{\tabcolsep}{0.12cm}
\scalebox{0.8}{
\begin{tabular}{lrrr:rr:r}
\hline
 & \multicolumn{1}{c}{Vocab.} & \multicolumn{1}{c}{Params} & \multicolumn{2}{c}{\textit{\textbf{OOV\_SD}}} & \multicolumn{2}{c}{\textit{\textbf{OOV\_RS}}} \\ 
 & \multicolumn{1}{c}{Size} & \multicolumn{1}{c}{Increment} & \multicolumn{1}{c:}{R-LCS} & \multicolumn{1}{c}{BSr} & \multicolumn{1}{c:}{R-LCS} & \multicolumn{1}{c}{BSr} \\ \hline

\multicolumn{7}{c}{\textbf{Medical}} \\
\textbf{23.92} & 75.43 \\
Llama-\textsc{VocabAdapt} w/o Replace & 141791 & 110M  & 26.56 & 76.52 & 23.86 & 75.54 \\
Llama-\textsc{VocabAdapt} w/ Replace & 140263 & 98M & \textbf{26.68} & \textbf{76.55} & 23.65 & \textbf{75.58} \\ \hdashline
Qwen-\textsc{VocabAdapt} w/o Replace & 162738 & 79M & 26.11 & 76.15 & \textbf{23.12} & \textbf{75.21} \\
Qwen-\textsc{VocabAdapt} w/ Replace & 158751 & 50M & \textbf{26.43} & \textbf{76.19} & 23.06 & 74.98 \\ \hline

\multicolumn{7}{c}{\textbf{Legal}} \\
Llama-\textsc{VocabAdapt} w/o Replace & 139470 & 91M & 24.17 & 67.45 & 23.71 & 67.64 \\
Llama-\textsc{VocabAdapt} w/ Replace & 137942 & 79M & \textbf{24.89} & \textbf{68.12} & \textbf{23.92} & \textbf{68.11} \\ \hdashline
Qwen-\textsc{VocabAdapt} w/o Replace & 162352 & 77M & 23.08 & 66.04 & 22.26 & 66.13 \\
Qwen-\textsc{VocabAdapt} w/ Replace & 158365 & 48M & \textbf{23.60} & \textbf{66.19} & \textbf{23.22} & \textbf{66.89} \\ \hline
\end{tabular}}
\caption{Ablation analysis for vocabulary adaptation methods with and without replacement. We show vocabulary sizes, parameter counts (in Millions), and performance metrics, R-LCS and BERTScore in challenging scenarios. We note that replacement-based methods (i) save $12.04\%$ parameters in Llama-3.1 and $37.19\%$ parameters in Qwen2.5-7B, (ii) performs better than without replacement in $13/16$ settings.}
\label{tab:with-wo-replace-stats}
\end{table}

\paragraph{Ablation analysis of vocabulary adaptation with and without replacement.} We report an ablation of vocabulary adaptation techniques with and without replacement in Table~\ref{tab:with-wo-replace-stats}. We note that replacement-based strategies perform better or at par with without-replacement strategies in 13 out of 16 settings. Replacement based strategies favors Llama (7 out of 8 settings) slightly more than Qwen (6 out of 8 settings). Contrary to previous discussions on higher OOV concentration in medical domain, we find here that Legal domain benefits more (all 8 settings) than medical domain (5 out of 8 settings). One possible explanation for this observation could be higher replacement fraction in legal domain ($25.43\%$) compared to medical domain ($23.81\%$). Importantly, replacement-based vocabulary adaptation does not increase the number of trainable parameters beyond the expanded embedding and unembedding layer ($lm\_head$). We note that replacement-based methods save $12.04\%$ parameters in Llama-3.1 and $37.19\%$ parameters in Qwen2.5-7B.

\paragraph{Comparison with closed-source LLMs.} We conducted a zero-shot analysis on a closed-source model: GPT-5 (\texttt{gpt-5-mini-2025-08-07}). We aim to understand that as the number of parameters increases, does over-fragmentation still persists as an underlying issue. To that end, we ran the evaluation on GPT-5 and compared the results with the best of Llama and Qwen results on \textsc{VocabAdapt} method. The results are shown in Table~\ref{tab:compare_gpt_5}.

\begin{table}[ht]
\centering
\scriptsize
\scalebox{0.8}{
\begin{tabular}{lcccccc}
\toprule
 & \multicolumn{2}{c}{Random} & \multicolumn{2}{c}{\textbf{\textit{OOV\_SD}}} & \multicolumn{2}{c}{\textbf{\textit{OOV\_RS}}} \\
\cmidrule(lr){2-3} \cmidrule(lr){4-5} \cmidrule(lr){6-7}
Model & R-LCS & BSr & R-LCS & BSr & R-LCS & BSr \\
\midrule
\multicolumn{7}{c}{\textbf{Medical}} \\
GPT-5-mini          & 21.80 & 73.89 & 23.95 & 75.07 & 22.65 & 74.81 \\
\textsc{VocabAdapt}\textsubscript{Best}   & \textbf{24.98} & \textbf{75.98} & \textbf{26.68} & \textbf{76.55} & \textbf{23.65} & \textbf{75.58} \\
\midrule
\multicolumn{7}{c}{\textbf{Legal}} \\
GPT-5-mini          & 19.94 & 67.54 & 19.45 & 66.40 & 20.46 & 67.08 \\
\textsc{VocabAdapt}\textsubscript{Best}   & \textbf{25.42} & \textbf{69.14} & \textbf{24.89} & \textbf{68.12} & \textbf{23.92} & \textbf{68.11} \\
\bottomrule
\end{tabular}}
\caption{Performance of GPT-5-mini and \textsc{VocabAdapt}\textsubscript{Best} on summarization across Medical and Legal domains, evaluated using Rouge-LCS (R-LCS) and BERTScore (BSr) under challenging scenarios (\textbf{\textit{OOV\_SD}} and \textbf{\textit{OOV\_RS}}) and Random setting.}
\label{tab:compare_gpt_5}
\end{table}

We note that our 7-8B parameter model with vocabulary adaptation is consistently outperforming gpt-5-mini (speculated several orders larger than 7B model with more complex architecture and workflow including MoE imbibed) in all the scenarios. This motivates the need for vocabulary adaptation even for larger parameter models.
\section{Related Works}
\paragraph{Domain Adaptation via Continued Pretraining.} Standard adaptation strategies rely on continued pretraining (CPT) to align generalist models to expert domain. Prominent examples include MEDITRON \citep{chen2023meditron70b} BioMistral \citep{labrak-etal-2024-biomistral} and ChatLaw~\cite{cui2024chatlawmultiagentcollaborativelegal}, which utilize massive domain corpora to enhance performance. However, these model-centric approaches are computationally-intensive and fail to address the underlying tokenization over-fragmentation~\cite{si2019enhancing}, leading to inefficient inference and context window erosion \citep{gu2021domain}.

\paragraph{Vocabulary Expansion Strategies.} To mitigate fragmentation, recent research has pivoted towards vocabulary expansion. \citet{hong-etal-2021-avocado} introduced AVocaDo to optimize vocabulary based on fragment scores~\cite{rust-etal-2021-good}, while Task-Adaptive Tokenization \citep{liu-etal-2023-task} leverages subword regularization to reduce sequence length. More targeted approaches like MEDVOC \citep{balde2024ijcai}, Gold Panning \citep{liu-etal-2024-gold}, HYPEROFA~\cite{ozeren-etal-2025-hyperofa}, AdaptiVocab~\cite{nakashadaptivocab}, and MEDVOC-LLM and ScafFix~\cite{balde2025evaluationllmsmedicaltext} focus on selecting high-value domain tokens, though these additive methods inevitably increase the model's parameter count and memory footprint.

\paragraph{Vocabulary Pruning and Replacement.} Addressing parameter efficiency, emerging works investigates pruning and token recycling. \citet{land-bartolo-2024-fishing} identified "glitch tokens" as undertrained vocabulary artifacts ripe for removal. Building on this, methods like Vocab Diet \citep{reif2025vocabdietreshapingvocabulary}, and COMPACT \citep{kong2025token} demonstrate that pruning unused tokens or replacing them with domain-specific terms can maintain performance~\cite{purason2025teachingoldtokenizersnew}. This establishes the basis for our replacement-based framework, which achieves adaptation with relatively less parameter growth as compared to expansion techniques.

\section{Conclusion}
We presented a systematic study of vocabulary adaptation for domain-specific summarization, focusing on when and why it improves LLMs performance. Across controlled settings: \textbf{\textit{OOV\_RS}} and \textbf{\textit{OOV\_SD}}, we showed that gains are governed by the severity and location of vocabulary mismatch. Vocabulary adapted models converge faster ($35-55\%$) than continual pretraining alone. Furthermore, vocabulary adaptation not only improves performance quantitatively (in terms of ROUGE-LCS and BERTScore) but also qualitatively (coherence, relevance, and faithfulness) as noted in our LLM-as-a-Judge evaluation. Replacement-based strategies remain parameter-efficient saving up to $37\%$ parameters and further improve robustness over expansion-only counterpart. These findings position tokenization as a design consideration for future domain adaptation works. We make our codebase pulicly available at \url{https://github.com/gb-kgp/VocabReplace-Then-Expand}. 


\section{Limitations}
Our work has the the following limitations. First, we built our \textit{fixed-size} 100M pretraining corpora inspired from prior art~\cite{beltagy-etal-2019-scibert,chen2023meditron70b,cite:InLegalBert}; however, there can be many other ways to come up with a much more fine-grained pretraining corpora. This can be an interesting future work to explore. Second, we note that LLMs considered, Llama-3.1 and Qwen2.5, have large vocabulary sizes (128K and 151K); still, there is a significant overlap in the vocabulary of these models. However, this in no way affects the findings of this work. It could indeed be interesting to explore the efficacy of these strategies of other varying vocabulary size models, like Microsoft-Phi~\cite{abdin2024phi} with a vocabulary size of 100K, and Gemma~\cite{gemma3technicalreport} series with a vocabulary size of 256K. Third, we fix the size of expansion vocabulary at 10K based on the natural frequency order which we found resulted in decent fragment scores mitigating over-fragmentation. We speculate there can be more nuanced ways to carefully select this 10K subset, and leave this as a potential future work to explore.

\section{Ethics Statement and Broader Impact}
The LLMs considered in this study, Llama and Qwen family, are general purpose LLMs. Although our techniques are showing promising improvements, they are in no way ready for a production ready deployment before ensuring proper saftery checks and balances. There still needs to be more dedicated research to investigate hallucination, correctness, and completeness of the response in real-world open-ended generation. 

\section*{Acknowledgments}
We thank the Ministry of Education, Govt of India, for supporting Gunjan Balde with Prime Minister Research Fellowship during his Ph.D. tenure. This research was partially funded by a Google Academic Research Award. We acknowledge National Supercomputing Mission (NSM) for providing computing resources of `PARAM Shakti’ at IIT Kharagpur, implemented by C-DAC and supported by the Ministry of Electronics and Information Technology (MeitY) and Department of Science and Technology (DST), Government of India.
\if{0}
can be used 
In this work, we focused on existing vocabulary adaptation techniques in the medical domain. Specifically, we highlighted two underlying issues: a) the target-task-specific nature of these techniques, and b) expansion strategies that lead to an average 5-8\% increment in parameters. We tackle both challenges by proposing parameter-efficient target-task-agnostic vocabulary adaptation strategies. We find that such task-agnostic adaptation outperforms task-specific by $15.87\%$ in high-OOV settings. Then we focused on vocabulary replacement strategies that first identified \textit{under-utilized} tokens in original LLM vocabulary--tokens that are undertrained, untrained, under-represented, unreachable--and then replaced them with the vocabulary to be added. These strategies improve over existing expansion strategies by a margin of $12.95\%$ in overall fine-grained settings. 
\fi
\bibliography{custom}

\appendix
\section{Experiments and Results Details}\label{sec:appendix_expt_results}

\subsection{Sampling ICL Demonstration}\label{sec:appendix_sample_ICL}
In order to sample the ICL demonstration from train set per test example, we use cosine similarity over embeddings obtained from sentence-transformers model variant of PubMedBERT~\footnote{\url{https://huggingface.co/pritamdeka/PubMedBERT-mnli-snli-scinli-scitail-mednli-stsb}} for medical domain. Due to extremely lengthy nature of documents in legal domain, we use standard bm25 model inspired from prior art~\cite{joshi-etal-2024-il} to get the closest training demonstration for the test point. We use bm25s library~\cite{bm25s} to setup the retriever.

\subsection{Baseline Models} \label{sec:appendix_methods_eval}
Here we describe the methods evaluated in this work:

\paragraph{xx-BASE.} These are the BASE LLMs variant (not the instruction tuned ones): Llama-3.1-8B-BASE and Qwen2.5-7B-BASE. They have not undergone any vocabulary modification and continual pretraining.

\paragraph{CPTOnly (No Vocab Adapt).} These are the variants of BASE LLMs models have not undergone any vocabulary modification but only standard continual pretraining over the domain-specific corpora.

\paragraph{\textsc{VocabAdapt} W/o Replace.} These are pure vocabulary expansion baselines without any replacement. We directly take non-overlapping (from BASE LLMs vocabulary)  top-10K vocabulary tokens learn from the domain-specififc tokenizers and add it to the model vocabulary. The expansion and addition procedure is similar to MEDVOC~\cite{balde2025evaluationllmsmedicaltext}, where for each vocabulary token to be added we iteratively add its subwords as obtained from the BASE LLM tokenizer.

\paragraph{\textsc{VocabAdapt} W/ Replace.} This is the replacement variant of \textit{\textsc{VocabAdapt} W/o Replace}. Here, we first replace tokens from the candidate replace set--$V_{\text{cand}}$(Eq.~\ref{eq:final_candidate}), then expand the BASE LLMs vocabulary with the remaining vocabulary tokens.

\paragraph{\textsc{VocabAdapt}\textsubscript{Refine} W/o Replace.} These are vocabulary expansion baselines without any replacement. Here, before selecting the top 10K tokens for expansion, we do a refinement steps of removing non-standard tokens--mixture of chars and numbers, chars and punctuation, numbers and punctuations--that might be inconsistent with BASE LLM tokenization, inspired from MEDVOC-LLM~\cite{balde2025evaluationllmsmedicaltext}. Post-refinement we take non-overlapping (from BASE LLMs vocabulary)  top-10K vocabulary tokens learn from the domain-specififc tokenizers and add it to the model vocabulary.

\begin{table}[t]
\centering
\scriptsize
\setlength{\tabcolsep}{0.1cm}
\scalebox{0.8}{
\begin{tabular}{c|c|cc|cc|cc|cc}
\hline
 & \multicolumn{1}{c|}{\textbf{Corpus}} & \multicolumn{2}{c|}{\textbf{SD Token Count}} & \multicolumn{2}{c|}{\textbf{RS Token Count}} & \multicolumn{2}{c|}{\textbf{SD OOV Conc.}} & \multicolumn{2}{c}{\textbf{RS OOV Conc.}} \\
 & \multicolumn{1}{c|}{\textbf{Size}} & \multicolumn{1}{c}{\textbf{Llama}} & \multicolumn{1}{c|}{\textbf{Qwen}} & \multicolumn{1}{c}{\textbf{Llama}} & \multicolumn{1}{c|}{\textbf{Qwen}} & \multicolumn{1}{c}{\textbf{Llama}} & \multicolumn{1}{c|}{\textbf{Qwen}} & \multicolumn{1}{c}{\textbf{Llama}} & \multicolumn{1}{c}{\textbf{Qwen}} \\ \hline
\multicolumn{10}{c}{\textbf{Evidence-Based Summarization}} \\
\textbf{Random} & 185 & 392 & 410 & 92 & 97 & 7.81 & 7.81 & 8.38 & 8.38 \\
\textbf{\textit{OOV\_SD}} & 185 & 346 & 354 & 75 & 79 & 14.77 & 14.77 & 10.76 & 10.76 \\
\textbf{\textit{OOV\_RS}} & 185 & 382 & 395 & 53 & 54 & 9.93 & 9.93 & 19.10 & 19.10 \\ \hline
\multicolumn{10}{c}{\textbf{Clinical Healthcare Query Summarization}} \\
\textbf{Random}& 150 & 79 & 80 & 13 & 13 & 8.32 & 8.33 & 11.07 & 11.07 \\
\textbf{\textit{OOV\_SD}} & 114 & 51 & 52 & 14 & 14 & 23.30 & 23.33 & 15.14 & 15.14 \\
\textbf{\textit{OOV\_RS}} & 143 & 80 & 80 & 15 & 15 & 12.11 & 12.12 & 27.74 & 27.74\\ \hline
\end{tabular}}
\caption{Dataset statistics of summarization datasets under \textbf{\textit{Random}}, \textbf{\textit{OOV\_RS}}, and \textbf{\textit{OOV\_SD}} settings, reporting mean token counts, OOV concentration (fraction of unigrams in text split more than once).} 
\label{appendix:tab_dataset_desc} 
\end{table}

\begin{table}[t]
    \centering
    \scriptsize
    \scalebox{0.9}{
    \begin{tabular}{p{7cm}}
        \hline
        \multicolumn{1}{c}{\textbf{Prompt structure}} \\ \hline
        \multicolumn{1}{c}{\textbf{Evidence-Based Summarization}} \\
            You are an expert medical professional. \\
            \#\#\#\# \\
            
            Summarize the given source document in the context of the input query in 100 words or less. Use the examples to guide word choice. \\
            
            Query 1: \{Train-Query\} \\
            
            Source Document 1: \{Train-Source-Document\} \\
            
            Query-Focused Summary 1: \{Train-Target-Summary\} \\

            \#\# \\
            
            Query 2: \{Test-Query\} \\
            
            Source Document 2: \{Test-Source-Document\} \\
            
            Query-Focused Summary 2:  \\ \hline
        
        \multicolumn{1}{c}{\textbf{Clinical Healthcare Query Summarization}} \\
            You are an expert medical professional. \\
            \#\#\#\# \\
            Summarize the given patient healthcare query into a concise single question of 10 words or less. Use the examples to guide word choice. \\

            Patient Health Query 1: \{Train-Patient-Query\} \\
            Summarized Question 1 : \{Train-Summarized-Question\} \\
            \#\# \\
            Patient Health Query 2: \{Test-Patient-Query\} \\
            Discharge Summary 2 : \\ \hline              
        
    \end{tabular}}
    \caption{The prompt structure used for prompting LLMs inspired from the prompt structure proposed in ClinSumm~\cite{van2024adapted}. Since we are using BASE LLMs there is no explicit segregation of system prompt and user prompt.}
    \label{appeendix:tab_prompt_structure}
\end{table}

\paragraph{\textsc{VocabAdapt}\textsubscript{Refine} W/ Replace.} This is the replace-then-expand variant of 
\textit{\textsc{VocabAdapt}\textsubscript{Refine} W/o Replace.}

\subsection{Results Trends}
We report the full results in Table~\ref{tab:appendix_full-results}. We now provide discussions as observed across challenging scenarios.

\begin{table*}[]
\centering
\scriptsize
\setlength{\tabcolsep}{0.15cm}
\scalebox{0.85}{
\begin{tabular}{l|r|r|r|r:r|r:r|r:r|r:r|r:r|r:r}
\hline
 & \multicolumn{1}{c|}{\textbf{Vocab}} & \multicolumn{1}{c|}{\textbf{Param}} & \multicolumn{1}{c|}{\textbf{Best}} & \multicolumn{4}{c|}{\textit{\textbf{Random}}} & \multicolumn{4}{c|}{\textit{\textbf{OOV\_SD}}} & \multicolumn{4}{c}{\textit{\textbf{OOV\_RS}}} \\ \cline{5-16}
 & \multicolumn{1}{c|}{\textbf{Size}} & \multicolumn{1}{c|}{\textbf{Incr.}} & \multicolumn{1}{c|}{\textbf{Ckpt.}} & \multicolumn{1}{c:}{FrSr\textsubscript{SD}} & \multicolumn{1}{c|}{FrSr\textsubscript{RS}} & \multicolumn{1}{c}{R-LCS} & \multicolumn{1}{c|}{BSr} & \multicolumn{1}{c:}{FrSr\textsubscript{SD}} & \multicolumn{1}{c|}{FrSr\textsubscript{RS}} & \multicolumn{1}{c:}{R-LCS} & \multicolumn{1}{c|}{BSr} & \multicolumn{1}{c:}{FrSr\textsubscript{SD}} & \multicolumn{1}{c|}{FrSr\textsubscript{RS}} & \multicolumn{1}{c:}{R-LCS} & \multicolumn{1}{c}{BSr} \\ \hline

 \multicolumn{16}{c}{\textbf{MEDICAL --ClinSumm}} \\
Llama-3.1-8B-BASE & 128256 & - & - & 1.16 & 1.16 & 23.03 & 70.66 & 1.25 & 1.27 & 24.39 & 71.35 & 1.20 & 1.34 & 21.33 & 70.41 \\
CPTOnly (No Vocab Adapt) & 128256 & - & 7500 & 1.16 & 1.16 & 24.89 & 75.55 & 1.25 & 1.25 & 26.29 & 76.22 & 1.20 & 1.34 & \textbf{23.92} & 75.43 \\
\textsc{VocabAdapt} W/o Replace. & 141779 & 110M & 3500 & 1.05 & 1.06 & 24.81 & 75.81 & 1.09 & 1.09 & 25.89 & 76.36 & 1.06 & 1.12 & 23.41 & 75.48 \\
\textsc{VocabAdapt} W/ Replace. & 140251 & 98M & 3500 & 1.05 & 1.06 & 24.13 & 75.36 & 1.09 & 1.09 & 25.45 & 75.83 & 1.06 & 1.12 & 22.81 & 75.02 \\
\textsc{VocabAdapt}\textsubscript{Refine} W/o Replace. & 141791 & 110M & 3500 & 1.05 & 1.07 & 24.58 & 75.66 & 1.09 & 1.09 & 26.56 & 76.52 & 1.06 & 1.12 & 23.86 & 75.54 \\
\textsc{VocabAdapt}\textsubscript{Refine} W/ Replace. & 140263 & 98M & 3500 & 1.05 & 1.07 & \textbf{24.98} & \textbf{75.98} & 1.09 & 1.09 & \textbf{26.68} & \textbf{76.55} & 1.06 & 1.12 & 23.65 & \textbf{75.58} \\ \hdashline

Qwen2.5-7B-BASE & 151665 & - & - & 1.19 & 1.23 & 15.51 & 43.35 & 1.28 & 1.29 & 14.15 & 37.44 & 1.24 & 1.36 & 12.23 & 35.22 \\
CPTOnly (No Vocab Adapt) & 151665 & - & 8000 & 1.19 & 1.23 & 24.73 & 75.32 & 1.28 & 1.29 & 25.96 & 75.90 & 1.24 & 1.36 & 22.41 & 74.44 \\
\textsc{VocabAdapt} W/o Replace. & 162738 & 79M & 3500 & 1.09 & 1.10 & \textbf{25.04} & \textbf{75.72} & 1.12 & 1.11 & 26.11 & 76.15 & 1.10 & 1.14 & \textbf{23.12} & \textbf{75.21} \\
\textsc{VocabAdapt} W/ Replace. & 158751 & 50M & 3500 & 1.09 & 1.10 & 24.67 & 75.41 & 1.12 & 1.11 & \textbf{26.43} & \textbf{76.19} & 1.10 & 1.14 & 23.06 & 74.98 \\
\textsc{VocabAdapt}\textsubscript{Refine} W/o Replace. & 162745 & 79M & 3500 & 1.08 & 1.09 & 24.22 & 75.26 & 1.12 & 1.11 & 25.76 & 75.84 & 1.10 & 1.14 & 22.40 & 74.77 \\
\textsc{VocabAdapt}\textsubscript{Refine} W/ Replace. & 158758 & 51M & 3500 & 1.08 & 1.09 & 24.68 & 75.51 & 1.12 & 1.11 & 26.21 & 76.12 & 1.10 & 1.14 & 22.81 & 74.93 \\ \hline

\multicolumn{16}{c}{\textbf{MEDICAL --EBM}} \\
Llama-3.1-8B-BASE & 128256 & - & - & 1.07 & 1.08 & 18.31 & 67.14 & 1.24 & 1.12 & 17.75 & \textbf{71.66} & 1.12 & 1.30 & 15.90 & 68.20 \\
CPTOnly (No Vocab Adapt) & 128256 & 0M & 7500 & 1.07 & 1.08 & 19.99 & 74.90 & 1.24 & 1.12 & \textbf{17.76} & 71.30 & 1.12 & 1.30 & \textbf{16.58} & \textbf{72.56} \\
\textsc{VocabAdapt} W/o Replace. & 141779 & 110M & 3500 & 1.01 & 1.01 & 20.18 & 72.40 & 1.09 & 1.02 & 17.12 & 69.10 & 1.03 & 1.12 & 16.44 & 72.21 \\
\textsc{VocabAdapt} W/ Replace. & 140251 & 98M & 3500 & 1.01 & 1.01 & 20.20 & 73.99 & 1.09 & 1.02 & 16.85 & 69.73 & 1.03 & 1.12 & 16.30 & 72.23 \\
\textsc{VocabAdapt}\textsubscript{Refine} W/o Replace. & 141791 & 110M & 3500 & 1.01 & 1.01 & \textbf{20.60} & 72.38 & 1.09 & 1.02 & 16.85 & 70.45 & 1.03 & 1.12 & 16.34 & 72.44 \\
\textsc{VocabAdapt}\textsubscript{Refine} W/ Replace. & 140263 & 98M & 3500 & 1.01 & 1.01 & 19.43 & \textbf{74.28} & 1.09 & 1.02 & 16.74 & 71.05 & 1.03 & 1.12 & 16.24 & 72.23 \\ \hdashline

Qwen2.5-7B & 151665 & - & - & 1.13 & 1.13 & 15.31 & 55.33 & 1.26 & 1.17 & 14.53 & 62.12 & 1.17 & 1.33 & 13.29 & 55.76 \\
CPTOnly (No Vocab Adapt) & 151665 & 0M & 8000 & 1.13 & 1.13 & 17.79 & 63.70 & 1.26 & 1.17 & 14.39 & 57.32 & 1.17 & 1.33 & 12.27 & 55.93 \\
\textsc{VocabAdapt} W/o Replace. & 162738 & 79M & 3500 & 1.05 & 1.05 & \textbf{20.06} & 73.13 & 1.12 & 1.07 & \textbf{16.58} & \textbf{71.44} & 1.07 & 1.14 & 15.08 & 70.10 \\
\textsc{VocabAdapt} W/ Replace. & 158751 & 50M & 3500 & 1.05 & 1.05 & 19.77 & \textbf{73.90} & 1.12 & 1.07 & 16.56 & 71.43 & 1.07 & 1.14 & \textbf{15.19} & \textbf{70.81} \\
\textsc{VocabAdapt}\textsubscript{Refine} W/o Replace. & 162745 & 79M & 3500 & 1.05 & 1.05 & 19.48 & 70.15 & 1.12 & 1.07 & 15.40 & 64.45 & 1.07 & 1.14 & 14.80 & 63.72 \\
\textsc{VocabAdapt}\textsubscript{Refine} W/ Replace. & 158758 & 51M & 3500 & 1.05 & 1.05 & 18.99 & 69.80 & 1.12 & 1.07 & 15.41 & 66.21 & 1.07 & 1.14 & 14.78 & 66.32 \\ \hline

\multicolumn{16}{c}{\textbf{MEDICAL --CHQ}} \\
Llama-3.1-8B-BASE & 128256 & - & - & 1.08 & 1.17 & \textbf{43.81} & 83.44 & 1.35 & 1.25 & 50.06 & 85.11 & 1.39 & 1.51 & \textbf{48.54} & 84.23 \\
CPTOnly (No Vocab Adapt) & 128256 & 0M & 7500 & 1.08 & 1.17 & 43.30 & 83.67 & 1.35 & 1.25 & 51.32 & 85.94 & 1.39 & 1.51 & 47.69 & \textbf{84.93} \\
\textsc{VocabAdapt} W/o Replace. & 141779 & 110M & 3500 & 1.06 & 1.08 & 43.69 & \textbf{83.98} & 1.28 & 1.13 & 50.37 & \textbf{86.03} & 1.09 & 1.27 & 46.88 & 84.52 \\
\textsc{VocabAdapt} W/ Replace. & 140251 & 98M & 3500 & 1.06 & 1.08 & 42.99 & 83.81 & 1.28 & 1.13 & \textbf{51.41} & 85.94 & 1.09 & 1.27 & 46.67 & 84.38 \\
\textsc{VocabAdapt}\textsubscript{Refine} W/o Replace. & 141791 & 110M & 3500 & 1.06 & 1.08 & 42.14 & 83.51 & 1.28 & 1.13 & 50.22 & 85.28 & 1.09 & 1.27 & 46.58 & 84.62 \\
\textsc{VocabAdapt}\textsubscript{Refine} W/ Replace. & 140263 & 98M & 3500 & 1.06 & 1.08 & 41.58 & 83.31 & 1.28 & 1.13 & 50.60 & 85.69 & 1.09 & 1.27 & 46.82 & 84.70 \\ \hdashline

Qwen2.5-7B & 151665 & - & - & 1.09 & 1.17 & 40.59 & 83.84 & 1.36 & 1.25 & 46.56 & 85.17 & 1.52 & 1.51 & 49.60 & 85.27 \\
CPTOnly (No Vocab Adapt) & 151665 & 0M & 8000 & 1.09 & 1.17 & \textbf{43.09} & \textbf{84.02} & 1.36 & 1.25 & 51.44 & 85.99 & 1.52 & 1.51 & 48.67 & 84.75 \\
\textsc{VocabAdapt} W/o Replace. & 162738 & 79M & 3500 & 1.07 & 1.08 & 42.09 & 83.55 & 1.29 & 1.13 & 49.22 & 85.90 & 1.09 & 1.27 & 48.08 & 84.99 \\
\textsc{VocabAdapt} W/ Replace. & 158751 & 50M & 3500 & 1.07 & 1.08 & 41.28 & 83.24 & 1.29 & 1.13 & 50.00 & 86.20 & 1.09 & 1.27 & 48.25 & 85.12 \\
\textsc{VocabAdapt}\textsubscript{Refine} W/o Replace. & 162745 & 79M & 3500 & 1.07 & 1.08 & 41.60 & 83.77 & 1.29 & 1.13 & 47.86 & 85.74 & 1.09 & 1.27 & 47.26 & 85.10 \\
\textsc{VocabAdapt}\textsubscript{Refine} W/ Replace. & 158758 & 51M & 3500 & 1.07 & 1.08 & 42.85 & 83.83 & 1.29 & 1.13 & 47.41 & 85.15 & 1.09 & 1.27 & 48.22 & 85.19 \\ \hline

\multicolumn{16}{c}{\textbf{LEGAL}} \\
Llama-3.1-8B-BASE & 128256 & - & - & 1.03 & 1.03 & \textbf{25.89} & 67.04 & 1.08 & 1.06 & 24.82 & 64.10 & 1.06 & 1.08 & \textbf{24.86} & 65.28 \\
CPTOnly (No Vocab Adapt) & 128256 & - & 10000 & 1.03 & 1.03 & 25.36 & 69.05 & 1.08 & 1.06 & 24.83 & 68.04 & 1.06 & 1.08 & 24.36 & \textbf{68.38} \\
\textsc{VocabAdapt} W/o Replace. & 139470 & 91M & 6500 & 1.01 & 1.01 & 25.47 & 69.05 & 1.01 & 1.01 & 24.17 & 67.45 & 1.01 & 1.01 & 23.71 & 67.64 \\
\textsc{VocabAdapt} W/ Replace. & 137942 & 79M & 6500 & 1.01 & 1.01 & 25.42 & \textbf{69.14} & 1.01 & 1.01 & \textbf{24.89} & \textbf{68.12} & 1.01 & 1.01 & 23.92 & 68.11 \\
\textsc{VocabAdapt}\textsubscript{Refine} W/o Replace. & 139653 & 93M & 6500 & 1.01 & 1.01 & 24.91 & 68.86 & 1.01 & 1.01 & 24.49 & 67.72 & 1.01 & 1.01 & 23.59 & 67.71 \\
\textsc{VocabAdapt}\textsubscript{Refine} W/ Replace. & 138125 & 81M & 6500 & 1.01 & 1.01 & 24.61 & 68.59 & 1.01 & 1.01 & 23.93 & 67.28 & 1.01 & 1.01 & 23.41 & 67.55 \\ \hdashline

Qwen2.5-7B-BASE & 151665 & - & - & 1.06 & 1.06 & 10.63 & 28.25 & 1.11 & 1.10 & 9.97 & 27.64 & 1.09 & 1.12 & 9.90 & 27.15 \\
CPTOnly (No Vocab Adapt) & 151665 & - & 10500 & 1.06 & 1.06 & \textbf{25.68} & \textbf{69.04} & 1.11 & 1.10 & \textbf{25.16} & \textbf{67.60} & 1.09 & 1.12 & \textbf{24.72} & \textbf{67.97} \\
\textsc{VocabAdapt} W/o Replace. & 162206 & 76M & 6500 & 1.02 & 1.02 & 23.31 & 66.83 & 1.05 & 1.05 & 22.22 & 65.25 & 1.04 & 1.05 & 21.95 & 65.67 \\
\textsc{VocabAdapt} W/ Replace. & 158219 & 47M & 6500 & 1.02 & 1.02 & 24.12 & 67.89 & 1.05 & 1.05 & 23.32 & 66.25 & 1.04 & 1.05 & 22.91 & 66.94 \\
\textsc{VocabAdapt}\textsubscript{Refine} W/o Replace. & 162352 & 77M & 6500 & 1.02 & 1.02 & 23.54 & 67.29 & 1.05 & 1.05 & 23.08 & 66.04 & 1.04 & 1.05 & 22.26 & 66.13 \\
\textsc{VocabAdapt}\textsubscript{Refine} W/ Replace. & 158365 & 48M & 6500 & 1.02 & 1.02 & 24.23 & 67.89 & 1.05 & 1.05 & 23.60 & 66.19 & 1.05 & 1.05 & 23.22 & 66.89 \\ \hline

\end{tabular}}
\caption{Comparison of vocabulary adaptation methods across different Legal and Medical domains using in-context learning with 1 ICL demonstration in two challenging scenarios--\textbf{\textit{OOV\_RS}} and \textbf{\textit{OOV\_SD}}, and a \textbf{\textit{Random}} subset. Performance is measured using Rouge-LCS (R-LCS) and BertScore (BSr) metrics.}
\label{tab:appendix_full-results}
\end{table*}

\paragraph{EBM and CHQ results.} The dataset statistics for these datasets is reported in Table ~\ref{appendix:tab_dataset_desc}. The prompt structure used for inference is reported in Table~\ref{appeendix:tab_prompt_structure}. We conduct evaluation in line with our challenge-oriented evaluation focusing on points that are difficult to generate (\textbf{\textit{OOV\_RS}}) and difficult to encode (\textbf{\textit{OOV\_SD}}). We note that \textsc{VOCABADAPT} outperforms BASE in a total of 11 out of 16 comparisons and CPTOnly in 9 out of 16 comparisons of difficult scenarios around \textbf{\textit{OOV\_RS}} and \textbf{\textit{OOV\_SD}} showing exactly the same behavior as observed in the paper. This further strengthens the generalizability of our approach and supports a broader task coverage.

\paragraph{\textbf{\textit{OOV\_SD}} (High OOV in Source Documents).}
In the \textbf{\textit{OOV\_SD}} setting, the test set is explicitly constructed from documents whose \emph{inputs} exhibit the highest OOV concentration, making accurate content understanding and alignment particularly challenging. Results show that BASE and CPTOnly models degrade noticeably in both R-LCS and BERTScore, indicating that continual pretraining alone is insufficient when the source text itself is dominated by unseen or poorly tokenized terms. Vocabulary adaptation methods consistently improve performance, demonstrating that better lexical coverage at the input level directly enhances content selection and factual grounding in summaries. Refinement-based expansion yields the most stable gains, suggesting that removing noisy or irregular candidate tokens before expansion helps the model form cleaner input representations. Replacement-based variants offer additional improvements in some cases, but the gains are less uniform, highlighting the sensitivity of source-side comprehension to overly aggressive vocabulary restructuring. Overall, \textbf{\textit{OOV\_SD}} emphasizes the importance of robust input tokenization, where accurate segmentation of domain-specific terms is critical for downstream summarization quality.

\paragraph{\textbf{\textit{OOV\_RS}} (High OOV in Reference Summaries).}
In contrast, \textbf{\textit{OOV\_RS}} focuses on datapoints where the \emph{references}—rather than the sources—contain high OOV concentration, stressing the model’s ability to generate or align with rare and domain-specific lexical forms. While BASE and CPTOnly models perform reasonably on the \textbf{\textit{Random}} split, they lag behind vocabulary-adapted models in \textbf{\textit{OOV\_RS}}, particularly in R-LCS, indicating difficulty in matching reference phrasing and terminology. Vocabulary expansion significantly narrows this gap, with consistent improvements across backbones, confirming that enhanced output-side lexical expressivity enables closer overlap with reference summaries. Refinement again proves beneficial by stabilizing gains across both metrics, whereas replacement-based methods yield modest but less consistent improvements. The contrast between \textbf{\textit{OOV\_SD}} and \textbf{\textit{OOV\_RS}} highlights that source-side OOV primarily affects content understanding, while reference-side OOV impacts lexical realization, and effective vocabulary adaptation is crucial in addressing both challenges beyond what is observed in the \textbf{\textit{Random}} setting.

\paragraph{Implications.}
Taken together, these results demonstrate that the benefits of vocabulary expansion scale with the severity and location of vocabulary mismatch. When OOVs are heavily concentrated in reference summaries (\textbf{\textit{OOV\_RS}}), vocabulary expansion directly improves generation fidelity. When OOVs originates in the source document (\textbf{\textit{OOV\_SD}}), vocabulary expansion becomes critical, yielding the largest and most consistent improvements. These findings highlight vocabulary mismatch as a bottleneck in expert domain adaptation and suggest that it should be selectively applied based on domain characteristics where vocabulary mismatch is indeed a significant problem. 

\begin{figure}[t]
    \centering
    \includegraphics[width=0.49\textwidth]{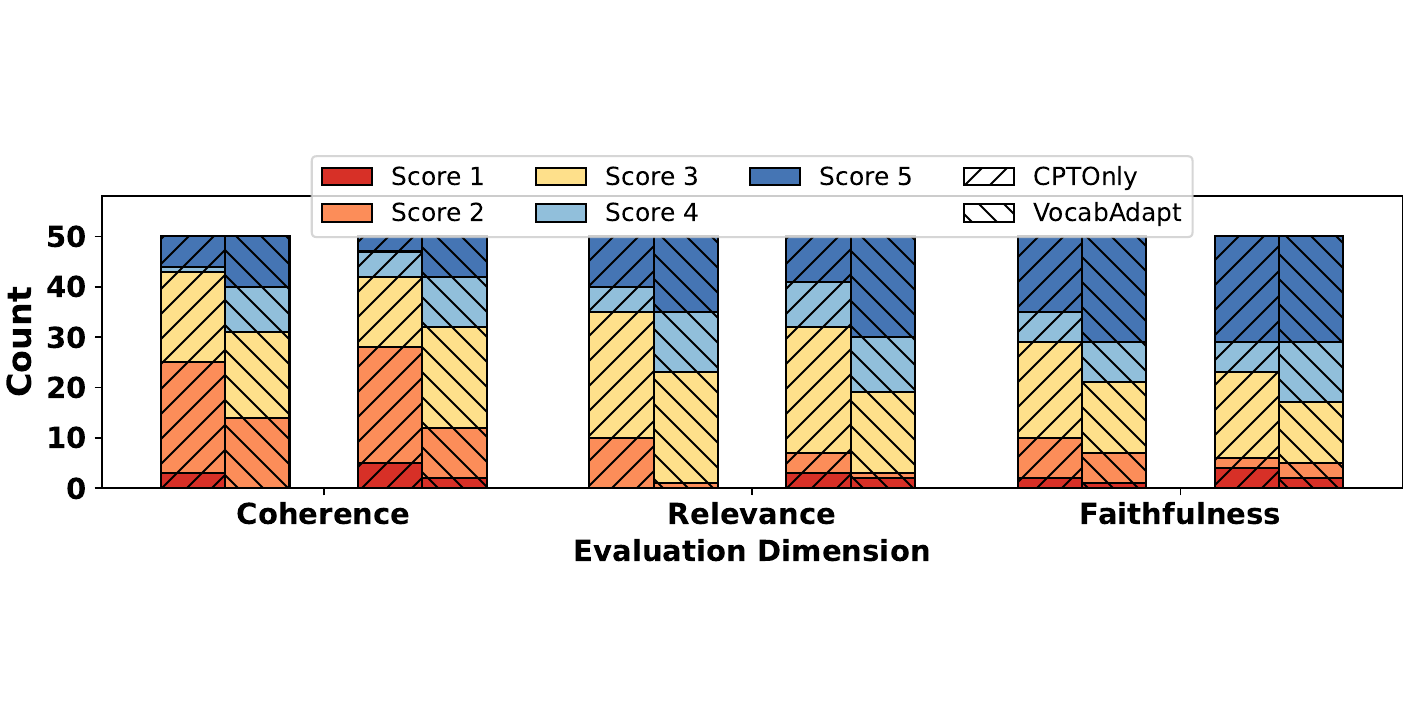}
    \caption{We report the score distribution as obtained from our LLM-as-a-Judge evaluation for medical domain. Acoss each dimension, the bars to the left corresponds to Llama model and bars to the right are for Qwen model. We note consistently \textsc{VocabAdapt} results in higher score of 4 or 5 compared to CPTOnly.}
    \label{fig:appendix_scores_LLaJ_medical}
\end{figure}

\begin{figure}[!ht]
    \centering
    \includegraphics[width=0.49\textwidth]{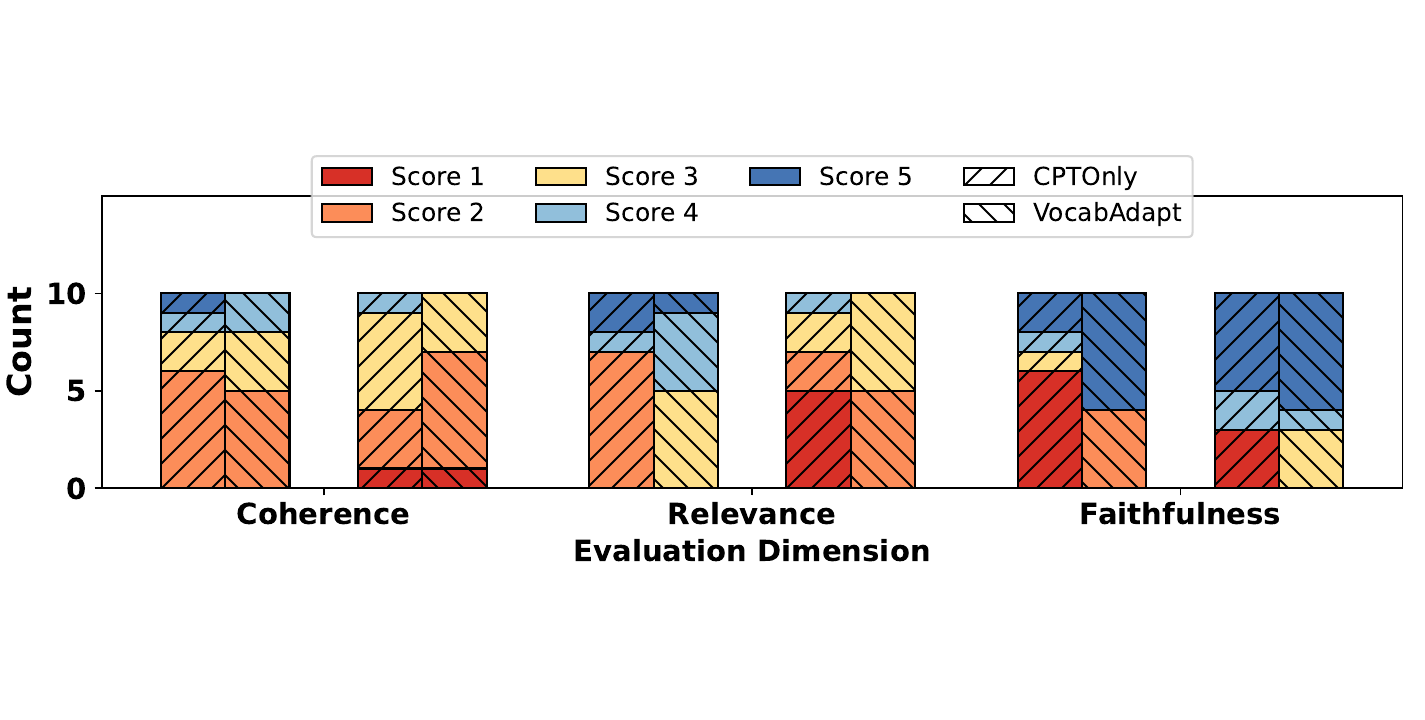}
    \caption{We report the score distribution as obtained from our LLM-as-a-Judge evaluation for Legal domain. Acoss each dimension, the bars to the left corresponds to Llama model and bars to the right are for Qwen model.}
    \label{fig:appendix_scores_LLaJ_legal}
\end{figure}

\subsection{LLM-as-a-Judge Evaluation}\label{sec:appendix_llaj}
We conduct LLM-as-a-judge (LlaJ) evaluation for summaries generated in medical and legal domain. We gather a uniform random subset of 100 summaries for medical domain and 20 summaries for legal domain from \textbf{\textit{OOV\_SD}} and \textbf{\textit{OOV\_RS}} settings distributed uniformly accorss Llama and Qwen models. We use models from Google's Gemma3 family~\cite{gemma3technicalreport}: MedGemma-27B-text-it\footnote{\url{https://huggingface.co/google/medgemma-27b-text-it}} model as our judge model for medical domain, and Gemma3-27B-it\footnote{\url{https://huggingface.co/google/gemma-3-27b-it}} as our judge model for Legal domain. The model is provided as input the source document and a generated summary. It is then asked to rate the generated summary on three dimensions in three separate runs: (i) coherence, (ii) relevance, and (iii) faithfulness; each on a scale of 1-5 (higher better). This is done for both \textsc{VocabAdapt} and CPTOnly summaries. The detailed system prompts and user prompts for each of the settings are available in the codebase inside  the folder "Random-Eval-LlaJ" folder. Our final scores are reported as average across across summary pairs in Table~\ref{tab:llaj-scores} and Figures~\ref{fig:appendix_scores_LLaJ_medical} and ~\ref{fig:appendix_scores_LLaJ_legal}. 

\subsection{Supplementary Human Evaluation}
\begin{figure}[t]
    \centering
    \includegraphics[width=0.95\columnwidth]{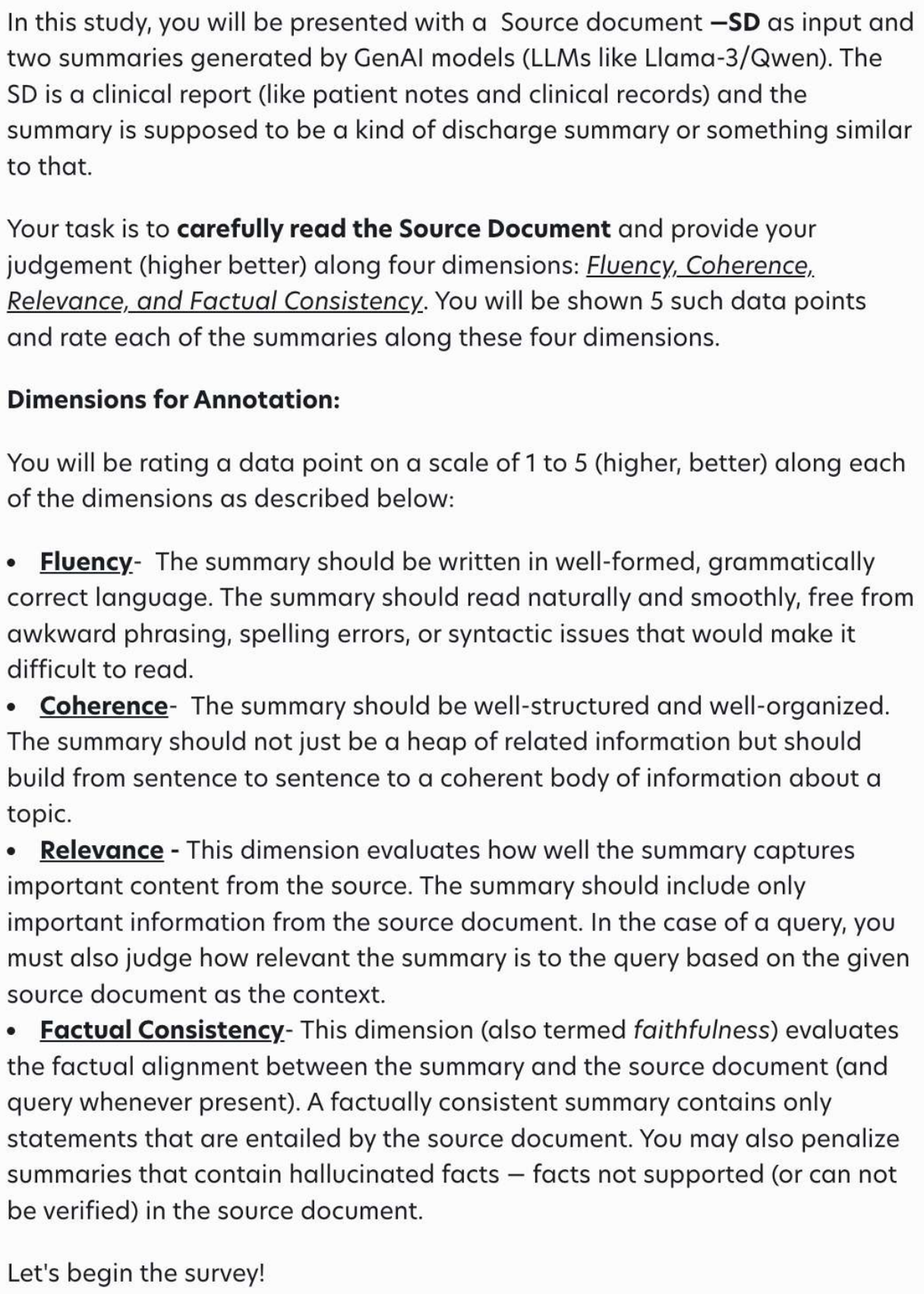}
    \caption{Annotation instructions shown to the participants.}
    \label{appendix:fig_annot_inst}
\end{figure}

We conducted an additional human assessment across fluency, consistency, relevance, and coherence rated on a scale of 1-5 on a subset of 20 summaries for the medical domain. The annotation instructions are shown in Figure~\ref{appendix:fig_annot_inst}. The evaluation was conducted on the Prolific platform with the following participation criterion:

\begin{itemize}
    \item \textbf{\textit{Highest education level completed}}. Undergraduate degree (BA/BS/other) \textit{OR} Graduate degree (MA/MS/MPhil/other) \textit{OR} Doctorate degree (PhD/other)
    \item \textbf{Employment Status.} Full-Time \textit{OR} Part-Time \textit{OR} Due to start a new job within the next month
    \item \textbf{Subject}: Biochemistry (Molecular and Cellular) \textit{OR} Biological Sciences \textit{OR} Biology \textit{OR} Biomedical Sciences
\end{itemize}

Each annotator was shown five summary pairs and each summary pair was evaluated independently by three annotators. The median time to complete the study was 20 minutes. In total 12 annotators were hired for the evaluation task. All the annotators were compensated at a rate of GBP 9 per hour. The average results across each category of annotation are shown in Table~\ref{appendix:tab_human_eval}.

\begin{table}[!ht]
    \centering
    \scriptsize
    \setlength{\tabcolsep}{0.1cm}
    \scalebox{0.8}{
    \begin{tabular}{ccccp{1.5cm}c}
    \hline
    Model & Fluency	& Coherence	& Relevance & Factual Consistency & Overall \\ \hline
    CPTOnly 	 & 4.12	 & 4.23	 & 4.17	 & 4.25	 & 4.19 \\
    \textsc{VOCABADAPT}	 & \textbf{4.35} & \textbf{4.40}	 & \textbf{4.32}	 & \textbf{4.57}	 & \textbf{4.41} \\ \hline
    
    \end{tabular}}
    \caption{Human evaluation trends comparing competing baseline and our propose \textsc{VocabAdapt} method.}
    \label{appendix:tab_human_eval}
\end{table}

The human evaluation exhibits the same overall trend as the LLM-as-a-Judge results; vocabulary adaptation models generate better summaries than the CPTOnly counterpart; reinforcing the validity of our conclusions. However, we must here mention conducting domain-specific human evaluation presented substantial practical challenges. Evaluating only 20 summary pairs in the medical domain incurred us a cost of approximately \textbf{GBP 48 via Prolific} (GBP 36 for annotators and GBP 12 as platform fees). \textbf{\textit{In the legal domain, Prolific does not even provide a sufficiently large or appropriate participant pool to enable reliable evaluation}}. These constraints make large-scale domain-expert evaluation difficult to sustain. 

We therefore view LLM-as-a-Judge not as a replacement for human evaluation, but as a scalable and reproducible alternative that is particularly valuable in settings where domain expertise is limited, costly, or difficult to source. Our results demonstrate that, when validated against human judgments, it provides consistent and reliable comparative assessment.

\end{document}